\documentclass[conference]{IEEEtran}
\usepackage{multirow}
\usepackage{array}
\usepackage{float}
\usepackage{comment}
\usepackage{color}

\ifCLASSINFOpdf
\else
\fi
\begin{document}
%
\title{Explaining Explanations: An Overview of Interpretability of Machine Learning}

\author{\IEEEauthorblockN{Leilani H. Gilpin, David Bau, Ben Z. Yuan, Ayesha Bajwa, Michael Specter and Lalana Kagal}
\IEEEauthorblockA{Computer Science and Artificial Intelligence Laboratory\\
Massachusetts Institute of Technology\\
Cambridge, MA 02139\\
\{lgilpin, davidbau, bzy, abajwa, specter, lkagal\}@ mit.edu}}


%


\maketitle

\begin{abstract}
There has recently been a surge of work in explanatory artificial intelligence (XAI).  This research area tackles the important problem that complex machines and algorithms often cannot provide insights into their behavior and thought processes.  XAI allows users and parts of the internal system to be more transparent, providing explanations of their decisions in some level of detail. These explanations are important to ensure algorithmic fairness, identify potential bias/problems in the training data, and to ensure that the algorithms perform as expected. However, explanations produced by these systems is neither standardized nor systematically assessed. In an effort to create best practices and identify open challenges, we describe foundational concepts of explainability and show how they can be used to classify existing literature. We discuss why current approaches to explanatory methods especially for deep neural networks are insufficient. Finally, based on our survey, we conclude with suggested future research directions for explanatory artificial intelligence.


%
%
%
%
\end{abstract}



%
\IEEEpeerreviewmaketitle

\section{Introduction}
As autonomous machines and black-box algorithms begin making decisions previously entrusted to humans, it becomes necessary for these mechanisms to explain themselves. Despite their success in a broad range of tasks including advertising, movie and book recommendations, and mortgage qualification, there is general mistrust about their results. In 2016, Angwin et al. \cite{angwin2016compas} analyzed  Correctional Offender Management Profiling for Alternative Sanctions (COMPAS), a widely used criminal risk assessment tool, and found that its predictions were unreliable and racially biased. Along with this, deep neural networks (DNNs) have been shown to be easily fooled into misclassifying inputs with no resemblance to the true category \cite{easily-fooled}.  Extending this observation, a number of techniques have been shown for changing a network's classification of any image to any target class by making imperceptible alterations to the pixels \cite{intriguing, moosavi2017universal, papernot2016limitations}.  Adversarial examples are not confined to images; natural language networks can also be fooled \cite{nlp-fool}.  Trojaning attacks have been demonstrated \cite{trojaning} in which inputs remain unchanged, but imperceptible changes are hidden in deep networks to cause them to make targeted errors.  While some defense methods have been developed, more attack methods have also emerged \cite{harnessing, madry2018towards, kraft-thesis, carlini2017adversarial}, and susceptibility to unintuitive errors remains a pervasive problem in DNNs.  The potential for such unexpected behavior and unintentional discrimination highlights the need for explanations.

As a first step towards creating explanation mechanisms, there is a new line of research in interpretability, loosely defined as the science of comprehending what a model did (or might have done).  Interpretable models and learning methods show great promise; examples include visual cues to find the ``focus'' of deep neural networks in image recognition and proxy methods to simplify the output of complex systems.  However, there is ample room for improvement, since identifying dominant classifiers and simplifying the problem space does not solve all possible problems associated with understanding opaque models.  

%
%
%
%

We take the stance that interpretability alone is insufficient.  In order for humans to trust black-box methods, we need \emph{explainability} -- models that are able to summarize the reasons for neural network behavior, gain the trust of users, or produce insights about the causes of their decisions.  While interpetability is a substantial first step, these mechanisms need to \emph{also} be complete, with the capacity to defend their actions, provide relevant responses to questions, and be audited.  Although interpretability and explainability have been used interchangeably, we argue there are important reasons to distinguish between them.  Explainable models are interpretable by default, but the reverse is not always true.  

Some existing deployed systems and regulations make the need for explanatory systems urgent and timely.  With impending regulations like the European Union's ``Right to Explanation'' \cite{eu}, calls for diversity and inclusion in AI systems \cite{critical-crawford}, findings that some automated systems may reinforce inequality and bias \cite{arvind}, and requirements for safe and secure AI in safety-critical tasks \cite{russell-robust}, there has been a recent explosion of interest in interpreting the representations and decisions of black-box models.  These models are everywhere, and the development of interpretable and explainable models is scattered throughout various disciplines.  Examples of general ``explainable systems'' include interpretable AI, explainable ML, causality, safe AI, computational social science, and automatic scientific discovery.  Further, research in explanations and their evaluation are found in machine learning,  human computer interaction (HCI), crowd sourcing, machine teaching, AI ethics, technology policy, and many other disciplines. This paper aims to broadly engage the greater machine learning community on the intersection of these topics, to set best practices, to define key concepts, and to propose evaluation criteria for standardizing explanatory systems often considered in isolation.



In this survey, we present a set of definitions, construct a taxonomy, and present best practices to start to standardize interpretability and explanatory work in AI.  We review a number of approaches towards explainable AI systems and provide a taxonomy of how one can think about diverse approaches towards explainability.  In Section 2, we define key terms including ``explanation", ``interpretability", and ``explainability".  We compare and contrast our definitions with those accepted in the literature.  In Section 3, we review some classical AI approaches (e.g., causal modeling, constraint reasoning, intelligent user interfaces, planning) but focus mainly on explainable models for deep learning.  We provide a summary of related work papers in Section 4, highlighting differences between definitions of key terms including ``explanation", ``interpretability", and ``explainability". In Section 5, we present a novel taxonomy that examines \emph{what is being explained} by these explanations.   We conclude with a discussion addressing open questions and recommend a path to the development and adoption of explainable methods for safety-critical or mission-critical applications.  

\section{Background and Foundational Concepts}
In this section, we provide background information about the key concepts of interpretability and explanability, and describe the meaningful differences between them.  

\subsection{What is an Explanation?}
Philosophical texts show much debate over what constitutes an explanation.  Of particular interest is what makes an explanation ``good enough'' or what really defines an explanation.  Some say a good explanation depends on the question \cite{bromberger}. This set of essays discusses the nature of explanation, theory, and the foundations of linguistics.  Although for our work, the most important and interesting work is on ``Why questions.'' In particular, when you can phrase what you want to know from an algorithm as a why question, there is a natural qualitative representation of when you have answered said question--when you can no longer keep asking why.  There are two why-questions of interest; why and why-should.  Similarly to the explainable planning literature, philosophers wonder about the why-shouldn't and why-should questions, which can give the kinds of explainability requirements we want.  

There is also discussion in philosophy about what makes the best explanation.  While many say it is inference \cite{thebest}, similar views point to the use of abductive reasoning to explain all the possible outcomes.  







\subsection{Interpretability vs. Completeness}


An explanation can be evaluated in two ways: according to its \textit{interpretability}, and according to its \textit{completeness}.

The goal of \emph{interpretability} is to describe the internals of a system in a way that is understandable to humans.  The success of this goal is tied to the cognition, knowledge, and biases of the user: for a system to be interpretable, it must produce descriptions that are simple enough for a person to understand using a vocabulary that is meaningful to the user.

The goal of \emph{completeness} is to describe the operation of a system in an accurate way.  An explanation is more complete when it allows the behavior of the system to be anticipated in more situations.  When explaining a self-contained computer program such as a deep neural network, a perfectly complete explanation can always be given by revealing all the mathematical operations and parameters in the system.

The challenge facing explainable AI is in creating explanations that are both complete and interpretable: it is difficult to achieve interpretability and completeness simultaneously.  The most accurate explanations are not easily interpretable to people; and conversely the most interpretable descriptions often do not provide predictive power.

Herman \cite{herman2017promise} notes that we should be wary of evaluating interpretable systems using merely human evaluations of interpretability, because human evaluations imply a strong and specific bias towards simpler descriptions.  He cautions that reliance on human evaluations can lead researchers to create \textit{persuasive} systems rather than transparent systems.  He presents the following ethical dilemmas that are a central concern when building interpretable systems:

\begin{center}
1) When is it unethical to manipulate an explanation \\ to better persuade users?

2) How do we balance our concerns for transparency and \\ ethics with our desire for interpretability?
\end{center}

We believe that it is fundamentally unethical to present a simplified description of a complex system in order to increase trust if the limitations of the simplified description cannot be understood by users, and worse if the explanation is optimized to hide undesirable attributes of the system. Such explanations are inherently misleading, and may result in the user justifiably making dangerous or unfounded conclusions.

To avoid this trap, explanations should allow a \textit{tradeoff} between interpretability and completeness.  Rather than providing only simple descriptions, systems should allow for descriptions with higher detail and completeness at the possible cost of interpretability.  Explanation methods should not be evaluated on a single point on this tradeoff, but according to how they behave on the curve from maximum interpretability to maximum completeness.

\subsection{Explainability of Deep Networks}
Explanations of the operation of deep networks have focused on either explaining the \textit{processing} of the data by a network, or explaining the \textit{representation} of data inside a network.  An explanation of processing answers ``Why does this particular input lead to that particular output?'' and is analogous to explaining the execution trace of a program.  An explanation about representation answers ``What information does the network contain?'' and can be compared to explaining the internal data structures of a program.

A third approach to interpretability is to create \textit{explanation-producing} systems with architectures that are designed to simplify interpretation of their own behavior. Such architectures can be designed to make either their processing, representations, or other aspects of their operation easier for people to understand.

\section{Review}
%
%

Due to the growing number of subfields, as well as the policy and legal ramifications \cite{eu} of opaque systems, the volume of research in interpretability is quickly expanding.  Since it is intractable to review all the papers in the space, we focus on explainable methods in deep neural architectures, and briefly highlight review papers from other subfields.  
\subsection{Explanations of Deep Network Processing}

Commonly used deep networks derive their decisions using a large number of elementary operations: for example, ResNet \cite{original-resnet}, a popular architecture for image classification, incorporates about $5\times 10^7$ learned parameters and executes about $10^{10}$ floating point operations to classify a single image.  Thus the fundamental problem facing explanations of such processing is to find ways to reduce the complexity of all these operations.  This can be done by either creating a \textit{proxy model} which behaves similarly to the original model, but in a way that is easier to explain, or by creating a \textit{salience map} to highlight a small portion of the computation which is most relevant.

\subsubsection{Linear Proxy Models}
The proxy model approach is exemplified well by the LIME method by Ribeiro \cite{why-trust}.  With LIME, a black-box system is explained by probing behavior on perturbations of an input, and then that data is used to construct a local linear model that serves as a simplified proxy for the full model in the neighborhood of the input. Ribeiro shows that the method can be used to identify regions of the input that are most influential for a decision across a variety of types of models and problem domains.  Proxy models such as LIME are predictive: the proxy can be run and evaluated according to its faithfulness to the original system. Proxy models can also be measured according to their model complexity, for example, number of nonzero dimensions in a LIME model. Because the proxy model  provides a quantifiable relationship between complexity and faithfulness, methods can be benchmarked against each other, making this approach attractive.

\subsubsection{Decision Trees}
Another appealing type of proxy model is the decision tree. Efforts to decompose neural networks into decision trees have recently extended work from the 1990s, which focused on shallow networks, to generalizing the process for deep neural networks.  One such method is DeepRED \cite{deepred}, which demonstrates a way of extending the CRED \cite{cred} algorithm (designed for shallow networks) to arbitrarily many hidden layers.  DeepRED utilizes several strategies to simplify its decision trees: it uses RxREN \cite{rxren} to prune unnecessary input, and it applies algorithm C4.5 \cite{quilan}, a statistical method for creating a parsimonious decision tree.  Although DeepRED is able to construct complete trees that are closely faithful to the original network, the generated trees can be quite large, and the implementation of the method takes substantial time and memory and is therefore limited in scalability.

Another decision tree method is ANN-DT \cite{ann-dt} which uses sampling to create a decision tree: the key idea is to use sampling to expand training using a nearest neighbor method.

\subsubsection{Automatic-Rule Extraction} 
Automatic rule extraction is another well-studied approach for summarizing decisions. Andrews et al \cite{rule-survey} outlines existing rule extraction techniques, and provides a useful taxonomy of five dimensions of rule-extraction methods including their expressive power, translucency and the quality of rules. Another useful survey can be found in the master's thesis by Zilke \cite{masters}.  

Decompositional approaches work on the neuron-level to extract rules to mimic the behavior of individual units.  The KT method \cite{fu} goes through each neuron, layer-by-layer and applies an if-then rule by finding a threshold.  Similar to DeepRED, there is a merging step which creates rules in terms of the inputs rather than the outputs of the preceding layer.  This is an exponential approach which is not tangible for deep neural networks.  However, a similar approach proposed by Tsukimoto \cite{tsukimoto} achieves polynomial-time complexity, and may be more tangible.  There has also been work on transforming neural network to fuzzy rules \cite{benitez}, by transforming each neuron into an approximate rule.  

Pedagogical approaches aim to extract rules by directly mapping inputs to outputs rather than considering the inner workings of a neural network.  These treat the network as a black box, and find trends and functions from the inputs to the outputs.  Validity Interval Analysis is a type of sensitivity analysis to mimic neural network behavior \cite{via}.  This method finds stable intervals, where there is some correlation between the input and the predicted class.  Another way to extract rules using sampling methods \cite{johansson, trepan}.  Some of these sampling approaches only work on binary input \cite{bio-re} or use genetic algorithms to produce new training examples \cite{kd-rule-ex}.  Other approaches aim to reverse engineer the neural network, notably, the RxREN algorithm, which is used in DeepRED\cite{deepred}.  

Other notable rule-extraction techniques include the MofN algorithm \cite{m-of-n}, which tries to find rules that explain single neurons by clustering and ignoring insignificant neurons.   Similarly, The FERNN \cite{fernn} algorithm uses the C4.5 algorithm \cite{quilan} and tries to identify the meaningful hidden neurons and inputs to a particular network.   

Although rule-extraction techniques increase the transparency of neural networks, they may not be truly faithful to the model.  With that, there are other methods that are focused on creating trust between the user and the model, even if the model is not ``sophisicated.''

\subsubsection{Salience Mapping}
The salience map approach is exemplified by occlusion procedure by Zeiler \cite{visualizing}, where a network is repeatedly tested with portions of the input occluded to create a map showing which parts of the data actually have influence on the network output.  When deep network parameters can be inspected directly, a salience map can be created more efficiently by directly computing the input gradient (Simonyan \cite{simonyan}).  Since such derivatives can miss important aspects of the information that flows through a network, a number of other approaches have been designed to propagate quantities other than gradients through the network.  Examples are LRP \cite{bach2015pixel}, DeepLIFT \cite{deep-lift}, CAM \cite{cam}, Grad-CAM \cite{grad-cam}, Integrated gradients \cite{integrated-gradients}, and SmoothGrad \cite{smooth-grad}.  Each technique strikes a balance between showing areas of high network activation, where neurons fire strongest, and areas of high network sensitivity, where changes would most affect the output.  A comparison of some of these methods can be found in Ancona \cite{ancona}.

\subsection{Explanations of Deep Network Representations}

While the number of individual operations in a network is vast, deep networks are internally organized into a smaller number of subcomponents: for example, the billions of operations of ResNet are organized into about 100 layers, each computing between 64 and 2048 channels of information per pixel.  The explanation of deep network representations aims to understand the role and structure of the data flowing through these bottlenecks.  This work can be divided by the granularity examined: representations can be understood \textit{by layer}, where all the information flowing through a layer is considered together, and \textit{by unit}, where single neurons or single filter channels are considered individually, and \textit{by vector}, where other vector directions in representation space are considered individually.

\subsubsection{Role of Layers}
Layers can be understood by testing their ability to help solve different problems from the problems the network was originally trained on.  For example Razavian \cite{razavian} found that the output of an internal layer of a network trained to classify images of objects in the ImageNet dataset produced a feature vector that could be directly reused to solve a number of other difficult image processing problems including fine-grained classification of different species of birds, classification of scene images, attribute detection, and object localization.  In each case, a simple model such as an SVM was able to directly apply the deep representation to the target problem, beating state-of-the-art performance without training a whole new deep network.  This method of using a layer from one network to solve a new problem is called \textit{transfer learning}, and it is of immense practical importance, allowing many new problems to be solved without developing new datasets and networks for each new problem.  Yosinksi \cite{yosinski} described a framework for quantifying transfer learning capabilities in other contexts. 

\subsubsection{Role of Individual Units}
The information within a layer can be further subdivided into individual neurons or individual convolutional filters.  The role of such individual units can be understood qualitatively, by creating visualizations of the input patterns that maximize the response of a single unit, or quantitatively, by testing the ability of a unit to solve a transfer problem.  Visualizations can be created by optimizing an input image using gradient descent \cite{simonyan}, by sampling images that maximize activation \cite{zhou2014object}, or by training a generative network to create such images \cite{nguyen-nips}.  Units can also be characterized quantitatively by testing their ability to solve a task.  One example of a such a method is network dissection \cite{netdissect2017}, which measures the ability of individual units solve a segmentation problem over a broad set of labeled visual concepts.  By quantifying the ability of individual units to locate emergent concepts such as objects, parts, textures, and colors that are not explicit in the original training set, network dissection can be used characterize the kind of information represented by visual networks at each unit of a network.

A review of explanatory methods focused on understanding unit representations used by visual CNNs can be found in \cite{visual}, which examines methods for visualization of CNN representations in intermediate network layers, diagnosis of these representations, disentanglement representation units, the creation of explainable models, and semantic middle-to-end learning via human-computer interaction.

Pruning of networks \cite{pruning} has also been shown to be a step towards understanding the role of individual neurons in networks.  In particular, large networks that train successfully contain small subnetworks with initializations conducive to optimization.  This demonstrates that there exist training strategies that make it possible to solve the same problems with much smaller networks that may be more interpretable. 

\subsubsection{Role of Representation Vectors}
Closely related to the approach of characterizing individual units is characterizing other directions in the representation vector space formed by linear combinations of individual units.  Concept Activation Vectors (CAVs) \cite{cavs} are a framework for interpretation of a neural net’s representations by identifying and probing directions that align with human-interpretable concepts.

\subsection{Explanation-Producing Systems}

Several different approaches can be taken to create networks that are designed to be easier to explain: networks can be trained to use \textit{explicit attention} as part of their architecture; they can be trained to learn \textit{disentangled representations}; or they can be directly trained to create \textit{generative explanations}.

\subsubsection{Attention Networks}
Attention-based networks learn functions that provide a weighting over inputs or internal features to steer the information visible to other parts of a network.  Attention-based approaches have shown remarkable success in solving problems such as allowing natural language translation models to process words in an appropriate non-sequential order \cite{vaswani2017attention}, and they have also been applied in domains such as fine-grained image classification \cite{xiao-two-level} and visual question answering \cite{hierarchical-co-attention}.  Although units that control attention are not trained for the purpose of creating human-readable explanations, they do directly reveal a map of which information passes through the network, which can serve as a form of explanation.  Datasets of human attention have been created \cite{human-VQA}, \cite{multimodal}; these allow systems to be evaluated according to how closely and their internal attention resembles human attention.

While attention can be observed as a way of extracting explanations, another interesting approach is to train attention explicitly in order to create a network that has behavior that conforms to desired explanations. This is the technique proposed by Ross \cite{ross2017right}, where input sensitivity of a network is adjusted and measured in order to create networks that are ``right for the right reasons;" the method can be used to steer the internal reasoning learned by a network. They also propose that the method can be used to learn a sequence of models that discover new ways to solve a problem that may not have been discovered by previous instances.

\subsubsection{Disentangled Representations}
Disentangled representations have individual dimensions that describe meaningful and independent factors of variation.  The problem of separating latent factors is an old problem that has previously been attacked using a variety of techniques such as Principal Component Analysis \cite{jolliffe1986principal}, Independent Component Analysis \cite{hyvarinen2000independent}, and Nonnegative Matrix Factorization \cite{berry2007algorithms}.  Deep networks can be trained to explicitly learn disentangled representations.  One approach that shows promise is Variational Autoencoding \cite{kingma2013auto}, which trains a network to optimize a model to match the input probability distribution according to information-theoretic measures.  Beta-VAE \cite{beta-vae} is a tuning of the method that has been observed to disentangle factors remarkably well.  Another approach is InfoGAN \cite{infogan}, which trains generative adversarial networks with an objective that reduces entanglement between latent factors.  Special loss functions have been suggested for encouraging feed-forward networks to also disentangle their units; this can be used to create interpretable CNNs that have individual units that detect coherent meaningful patches instead of difficult-to-interpret mixtures of patterns \cite{zhang2018interpretable}. Disentangled units can enable the construction of graphs \cite{zhang2} and decision trees \cite{zhang2018unsupervised} to elucidate the reasoning of a network. Architectural alternatives such as capsule networks \cite{capsule} can also organize the information in a network into pieces that disentangle and represent higher-level concepts.

\subsubsection{Generated Explanations}
Finally, deep networks can also be designed to generate their own human-understandable explanations as part of the explicit training of the system.  Explanation generation has been demonstrated as part of systems for visual question answering \cite{original-vqa} as well as in fine-grained image classification \cite{hendricks2016generating}.  In addition to solving their primary task, these systems synthesize a ``because'' sentence that explains the decision in natural language.  The generators for these explanations are trained on large data sets of human-written explanations, and they explain decisions using language that a person would use.

Multimodal explanations that incorporate both visual pointing and textual explanations can be generated; this is the approach taken in \cite{multimodal}.  This system builds upon the winner of the 2016 VQA challenge \cite{fukui}, with several simplification and additions.  In addition to the question answering task and the internal attention map, the system trains an additional long-form explanation generator together with a second attention map optimized as a visual pointing explanation.  Both visual and textual explanations score well individually and together on evaluations of user trust and explanation quality.  Interestingly, the generation of these highly readable explanations is conditioned on the output of the network: the explanations are generated based on the decision, after the decision of the network has already been made.


\section{Related Work}
We provide a summary of related review papers, and an overview of interpretability and explainability in other domains.

\subsection{Interpretability}

A previous survey has attempted to define taxonomies and best practices for a ``strong science'' of interpretability \cite{finale}.  The motivation of this paper is similar to ours, noting that ``the volume of research on interpretability is rapidly growing'' and that there is no clear existing definition or evaluation criteria for interpretability.  The authors define interpretability as ``the ability to explain or to present in understandable terms to a human'' and suggest a variety of definitions for explainability, converging on the notion that interpretation is the act of discovering the evaluations of an explanation.  The authors attempt to reach consensus on the definition of interpretable machine learning and how it should be measured. While we are inspired by the taxonomy of this paper, we focus on the explainability aspect rather than interpretability.  



The main contribution of this paper is a taxonomy of modes for interpretability evaluations: application-grounded, human-grounded, and functionally grounded. The authors state interpretability is required when a problem formulation is incomplete, when the optimization problem -- the key definition to solve the majority of machine learning problems -- is disconnected from evaluation. Since their problem statement is the incompleteness criteria of models, resulting in a disconnect between the user and the optimization problem, evaluation approaches are key.

The first evaluation approach is application-grounded, involving real humans on real tasks.  This evaluation measures how well human-generated explanations can aid other humans in particular tasks, with explanation quality assessed in the true context of the explanation's end tasks.  For instance, a doctor should evaluate diagnosis systems in medicine.

The second evaluation approach is human-grounded, using human evaluation metrics on simplified tasks. The key motivation is the difficulty of finding target communities for application testing. Human-grounded approaches may also be used when specific end-goals, such as identifying errors in safety-critical tasks, are not possible to realize fully. 

The final evaluation metric is functionally grounded evaluation, without human subjects.  In this experimental setup, proxy or simplified tasks are used to prove some formal definition of interpretability.  The authors acknowledge that choosing which proxy to use is a challenge inherent to this approach.  There lies a delicate tradeoff between choosing an interpretable model and a less interpretable proxy method which is more representative of model behavior; the authors acknowledge this point and briefly mention decision trees as a highly interpretable model.

The authors then discuss open problems, best practices and future work in interpretability research, while heavily encouraging data-driven approaches for discovery in interpretability. Although the contribution of the interpretability definition, we distinguish our taxonomy by defining  different focuses of explanations a model can provide, and how those explanations should be evaluated.  

\subsection{Explainable AI for HCI}
One previous review paper of explainable AI performed a sizable data-driven literature analysis of explainable systems \cite{trends}.  In this work, the authors move beyond the classical AI interpretability argument, focusing instead on how to create practical systems with efficacy for real users.  The authors motivate AI systems that are ``explainable by design'' and present their findings with three contributions: a data-driven network analysis of 289 core papers and 12,412 citing papers for an overview of explainable AI research, a perspective on trends using network analysis, and a proposal for best practices and future work in HCI research pertaining to explainablity.


Since most of the paper focuses on the literature analysis, the authors highlight only three large areas in their related work section: explainable artificial intelligence (XAI), intelligibility and interpretability in HCI, and analysis methods for trends in research topics.

The major contribution of this paper is a sizable literature analysis of explainable research, enabled by the citation network the authors constructed.  Papers were aggregated based on a keyword search on variations of the terms ``intelligible,'' ``interpretable,'' ``transparency,'' ``glass box,'' ``black box,'' ``scrutable,'' ``counterfacutals,'' and ``explainable,'' and then pruned down to 289 core papers and 12,412 citing papers.  Using network analysis, the authors identified 28 significant clusters and 9 distinct research communities, including early artificial intelligence, intelligent systems/agents/user interfaces, ambient intelligence, interaction design and learnability, interpretable ML and classifier explainers, algorithmic fairness/accountability/transparency/policy/journalism, causality, psychological theories of explanations, and cognitive tutors.  In contrast, our work is focused on the research in interpretable ML and classifier explainers for deep learning.

With the same sets of core and citing papers, the authors performed LDA-based topic modeling on the abstract text to determine which communities are related.  The authors found the largest, most central and well-studied network to be intelligence and ambient systems.  In our research, the most important subnetworks are the Explainable AI: Fair, Accountable, and Transparent (FAT) algorithms and Interpretable Machine Learning (iML) subnetwork and the theories of explanations subnetworks.  In particular, the authors provide a distinction between FATML and interpretability; while FATML is focused on societal issues,  interpretability is focused on methods.   Theory of explanations joins  causality and cognitive psychology with the common threads of counterfactual reasoning and causal explanations.   Both these threads are important factors in our taxonomy analysis.  
%
%


In the final section of their paper,  the authors name two trends of particular interest to us: ML production rules and a road map to rigorous and usable intelligibility.  The authors note a lack of classical AI methods being applied to interpretability, encouraging broader application of those methods to current research. 
Though this paper  focused mainly on setting an HCI research agenda in explainability, it raises many  points relevant to our work.  Notably, the literature analysis discovered subtopics and subdisciplines in psychology and social science, not yet identified as related in our analysis.  
\subsection{Explanations for Black-Box Models}
A recent survey on methods for explaining black-box models \cite{black-box-survey} outlined a taxonomy to provide classifications of the main problems with opaque algorithms.  Most of the methods surveyed are applied to neural-network based algorithms, and therefore related to our work.  

The authors provide an overview of methods that explaining decision systems based on opaque and obscure machine learning models. Their taxonomy is detailed, distinguishing small differing components in explanation approaches (e.g. Decision tree vs. single tree, neuron activation, SVM, etc.)  Their classification examines four features for each explanation method:
\begin{enumerate}
	\item The type of the problem faced.
    \item The explanatory capability used to open the black box.
    \item The type of black box model that can be explained.
    \item The type of input data provided to the black box model.
\end{enumerate}
They primarily divide the explanation methods according to the types of problem faced, and identify four groups of explanation methods: methods to explain black box models; methods to explain black box outcomes; methods to inspect black boxes; and methods to design transparent boxes.  Using their classification features and these problem definitions, they discuss and further categorize methods according to the type of explanatory capability adopted, the black box model ``opened'', and the input data.  Their goal is to review and classify the main black box explanation architectures, so their classifications can serve as a guide to identifying similar problems and approaches.  We find this work a meaningful contribution that is useful for exploring the design space of explanation methods.  Our classification is less finely-divided; rather than subdividing implementation techniques, we examine the focus of the explanatory capability and what each approach \emph{can} explain, with an emphasis on understanding how different types of explainability methods can be evaluated.

\subsection{Explainability in Other Domains}
Explainable planning \cite{planning} is an emerging discipline that exploits the model-based representations that exist in the planning community.  Some of the key ideas were proposed years ago in plan recognition \cite{generalized}.  Explainable planning urges the familiar and common basis for communication with users, while acknowledging the gap between planning algorithms and human problem-solving.  In this paper, the authors outline and provide examples of a number of different types of questions that explanations could answer, like ``Why did you do A'' or ''Why DIDN'T you do B'', ''Why CAN'T you do C'', etc.  In addition, the authors emphasize that articulating a plan in natural language is NOT usually the same thing as explaining the plan.
A request for explanation is ``an attempt to uncover a piece of knowledge that the questioner believes must be available to the system and that the questioner does not have''.  We discuss the questions an explanation can and should answer in our conclusion.

Automatic explanation generation is also closely related to computers and machines that can tell stories.  In John Reeves' thesis \cite{reeves}, he created the THUNDER program to read stories, construct character summaries, infer beliefs, and understand conflict and resolution.  Other work examines how to represent the necessary structures to do story understanding \cite{mueller-story}.  The Genesis Story-Understanding System \cite{genesis} is a working system that understands, uses, and composes stories using higher-level concept patterns and commonsense rules.  Explanation rules are used to supply missing causal or logical connections.  

At the intersection of human robot interaction and story-telling is verbalization; generating explanations for human-robot interaction \cite{verbalization}.   Similar approaches are found in abductive reasoning; using a case-based model \cite{abductive-focusing} or explanatory coherence \cite{coherence}.  This is also a well-studied field in brain and cognitive science by filling in the gaps of knowledge by imagining new ideas \cite{imagination} or using statistical approaches \cite{neural-prediction}.

\section{Taxonomy}

The approaches from the literature that we have examined fall into three different categories. Some papers propose explanations that, while admittedly non-representative of the underlying decision processes, provide some degree of \textit{justification} for emitted choices that may be used as response to demands for explanation in order to build human trust in the system's accuracy and reasonableness. These systems \textit{emulate} the \textit{processing} of the data to draw connections between the inputs and outputs of the system.  

The second purpose of an explanation is to explain the \textit{representation} of data inside the network.  These provide insight about the internal operation of the network and can be used to facilitate explanations or interpretations of activation data within a network.  This is comparative to explaining the internal data structures of the program, to start to gain insights about why certain intermediate representations provide information that enables specific choices.

The final type of explanation is \textit{explanation-producing} networks.  These networks are specifically built to explain themselves, and they are designed to simplify the interpretation of an opaque subsystem.  They are steps towards improving the transparency of these subsystems; where processing, representations, or other parts are justified and easier to understand.


The taxonomy we present is useful given the broad set of existing approaches for achieving varying degrees of interpretability and completeness in machine learning systems. Two distinct methods claiming to address the same overall problem may, in fact, be answering very different questions.  Our taxonomy attempts to subdivide the problem space, based on existing approaches, to more precisely categorize what has already been accomplished.


We show the classifications of our reviewed methods per category in Table \ref{tab:taxonomy}.  Notice that the processing and explanation-producing roles are much more populated than the representation role.  We believe that this disparity is largely due to the fact that it is difficult to evaluate representation-based models.  User-study evaluations are not always appropriate.  Other numerical methods, like demonstrating better performance by adding or removing representations, are difficult to facilitate.

The position of our taxonomy is to promote research and evaluation across categories.  Instead of other explanatory and interpretability taxonomies that assess the purpose of explanations \cite{finale} and their connection to the user \cite{trends}, we instead assess the focus on the method, whether the method tries to explain the \textit{processing} of the data by a network, explain the \textit{representation} of data inside a network or to be a self-explaining architecture to gain additional meta predictions and insights about the method.

We promote this taxonomy, particularly the explanation-producing sub-category, as a way to consider designing neural network architectures and systems.  We also highlight the lack of standardized evaluation metrics, and propose research crossing areas of the taxonomy as future research directions.  







\begin{table}[ht!]
    \centering
    \setlength{\extrarowheight}{1.5pt}
    \begin{tabular}{|c|c|c|}
        \hline
        \multirow{2}{*}{\bf Processing} & \multirow{2}{*}{\bf Representation} & {\bf Explanation} \\
        & &  {\bf Producing} \\
        \hline
        Proxy Methods & Role of layers & Scripted conversations\\
        Decision Trees & Role of neurons & Attention-based\\
        Salience mapping & Role of vectors & Disentangled rep.\\
        Automatic-rule extraction & & Human evaluation\\
         \hline 
    \end{tabular}
    \vspace{0.2cm}
    \caption{The classifications of top level methods into our taxonomy.}
    \label{tab:taxonomy}
\end{table}

\section{Evaluation}
Although we outline three different focuses of explanations for deep networks, they do not share the same evaluation criteria.  Most of the work surveyed conducts one of the following types of evaluation of their explanations.
\begin{enumerate}
\item Completeness compared to the original model.  A proxy model can be evaluated directly according to how closely it approximates the original model being explained.
\item Completeness as measured on a substitute task.  Some explanations do not directly explain a model's decisions, but rather some other attribute that can be evaluated.  For example, a salience explanation that is intended to reveal model sensitivity can be evaluated against a brute-force measurement of the model sensitivity.
\item Ability to detect models with biases.  An explanation that reveals sensitivity to a specific phenomenon (such as a presence of a specific pattern in the input) can be tested for its ability to reveal models with the presence or absence of a relevant bias (such as reliance or ignorance of the specific pattern).
\item Human evaluation.  Humans can evaluate explanations for reasonableness, that is how well an explanation matches human expectations.  Human evaluation can also evaluate completeness or substitute-task completeness from the point of view of enabling a person to predict behavior of the original model; or according to helpfulness in revealing model biases to a person.
\end{enumerate}


As we can see in Table \ref{tab:eval}, the tradeoff between \emph{interpretability} and its \emph{completeness} can be seen not only as a balance between simplicity and accuracy in a proxy model.  The tradeoff can also be made by anchoring explanations to substitute tasks or evaluating explanations in terms of their ability to surface important model biases. Each of the three types of explanation methods can provide explanations that can be evaluated for completeness (on those critical model characteristics), while still being easier to interpret than a full accounting for every detailed decision of the model.


\begin{table}[H]
    \centering
    \setlength{\extrarowheight}{1.5pt}
    \begin{tabular}{|c|c|c|}
        \hline
        \multirow{2}{*}{\bf Processing} & \multirow{2}{*}{\bf Representation} & {\bf Explanation} \\
        & &  {\bf Producing} \\
        \hline
        Completeness to Model & Completeness on & \multirow{2}{*}{Human evaluation} \\
        Completeness on &   substitute task & \multirow{2}{*}{Detect biases} \\
        substitute task &   Detect biases &  \\
         \hline 
    \end{tabular}
    \vspace{0.2cm}
    \caption{Suggested evaluations for the classifications in our taxonomy}
    \label{tab:eval}
\end{table}
\subsection{Processing}
Processing models can also be regarded as emulation-based methods.  Proxy methods should be evaluated on their faithfulness to the original model.  A handful of these metrics are described in \cite{why-trust}. The key idea is that evaluating completeness to a model should be local.  Even if a model, in our case, a deep neural network, is too complex globally, you can still explain in a way that makes sense locally by approximating local behavior.  Therefore, processing model explanations want to minimize the ``complexity'' of explanations (essentially, minimize length) as well as ``local completeness'' (error of interpretable representation relative to actual classifier, near instance being explained).

Salience methods that highlight sensitive regions for processing are often evaluated qualitatively.  Although they do not directly predict the output of the original method, these methods can also be evaluated for faithfulness, since their intent is to explain model sensitivity. For example, \cite{ancona} conducts an occlusion experiment as ground truth, in the model is tested on many version of an input image where each portion of the image is occluded.  This test determines in a brute-force but computationally inefficient way which parts of an input cause a model to change its outputs the most.  Then each salience method can be evaluated according to how closely the method produces salience maps that correlate with this occlusion-based sensitivity.

\subsection{Representation}
Representation-based methods typically characterize the role of portions of the representation by testing the representations on a transfer task.  For example, representation layers are characterized according to their ability to serve as feature input for a transfer problem, and both Network Dissection representation units and Concept Activation Vectors are measured according to their ability to detect or correlate with specific human-understandable concepts.

Once individual portions of a representation are characterized, they can be tested for explanatory power by evaluating whether their activations can faithfully reveal a specific bias in a network.  For example, Concept Activation Vectors \cite{cavs} are evaluated by training several versions of the same network on datasets that are synthesized to contain two different types of signals that can be used to determine the class (the image itself, and an overlaid piece of text which gives the class name with varying reliability).  The faithfulness of CAVs to the network behavior can be verified by evaluating whether classifiers that are known to depend on the text (as evidenced by performance on synthesized tests) exhibit high activations of CAV vectors corresponding to the text, and that classifiers that do not depend on the text exhibits low CAV vectors.

\subsection{Explanation-Producing}
Explanation-producing systems can be evaluated according to how well they match user expectations.  For example, network attention can be compared to human attention \cite{human-VQA}, and disentangled representations can be tested on synthetic datasets that have known latent variables, to determine whether those variables are recovered.  Finally, systems that are trained explicitly to generate human-readable explanations can be tested by similarity to test sets, or by human evaluation.

One of the difficulties of evaluating explanatory power of explanation-producing systems is that, since the system itself produces the explanation, evaluations necessarily couple evaluation of the system along with evaluation of the explanation.  An explanation that seems unreasonable could indicate either a failure of the system to process information in a reasonable way, or it could indicate the failure of the explanation generator to create a reasonable description.  Conversely, an explanation system that is not faithful to the decisionmaking process could produce a reasonable description even if the underlying system is using unreasonable rules to make the decision.  An evaluation of explanations based on their reasonableness alone can miss these distinctions.  In \cite{finale}, a number of user-study designs are outlined that can help bridge the gap between the model and the user.

\section{Conclusions}

One common viewpoint in the deep neural network community is that the level of interpretability and theoretical understanding needed to for transparent explanations of large DNNs remains out of reach; for example, as a response to Ali Rahimi's Test of Time NIPS address, Yann LeCunn responded that ``The engineering artifacts have almost always preceded the theoretical understanding'' \cite{lecun-response}.
However, we assert that, for machine learning systems to achieve wider acceptance among a skeptical populace, it is crucial that such systems be able to provide or permit satisfactory explanations of their decisions. The progress made so far has been promising, with efforts in explanation of deep network processing, explanation of deep network representation, and system-level explanation production yielding encouraging results.


We find, though, that the various approaches taken to address different facets of explainability are siloed. Work in the explainability space tends to advance a particular category of technique, with comparatively little attention given to approaches that merge different categories of techniques to achieve more effective explanation. Given the purpose and type of explanation, it is not obvious what the best type of explanation metric is and should be.  We encourage the use of diverse metrics that align with the purpose and completeness of the targeted explanation.  Our view is that, as the community learns to advance its work collaboratively by combining ideas from different fields, the overall state of system explanation will improve dramatically, resulting in methods that provide behavioral extrapolation, build trust in deep learning systems, and provide usable insight into deep network operation enabling system behavior understanding and improvement.

\section*{Acknowledgements}
The work was partially funded by DARPA XAI program FA8750-18-C0004, the National Science Foundation under Grants No. 1524817, the MIT-IBM Watson AI Lab, and the Toyota Research Institute (TRI).  The authors also wish to express their appreciation for Jonathan Frankle for sharing his insightful feedback on earlier versions of the manuscript.

\bibliographystyle{IEEEtran}
\bibliography{draft.bib}

\end{document}